\pdfoutput=1
\documentclass[11pt]{article}

\usepackage[final]{acl}

\usepackage{times}
\usepackage{latexsym}

\usepackage[T1]{fontenc}
\usepackage[utf8]{inputenc}

\usepackage{microtype}

\usepackage{inconsolata}

\usepackage{graphicx}

\usepackage{enumitem}
\setlist[enumerate]{itemsep=0pt, topsep=0pt}
\usepackage{booktabs, multirow} % for borders and merged ranges
\usepackage{soul}% for underlines
\usepackage{xcolor,colortbl} % for cell 
\usepackage[dvipsnames]{xcolor}
\usepackage{changepage,threeparttable} % for wide tables
\usepackage{amsmath}
\usepackage{float} % in preamble
\usepackage{subcaption}
\usepackage{arydshln} % in preamble
\usepackage{placeins}
\usepackage{amssymb}
\usepackage{svg}
\usepackage{subcaption} % put this in your preamble

\title{FormGym: Doing Paperwork with Agents}

\author{
 \textbf{Matthew Toles\textsuperscript{1}},
 \textbf{Isaac Song\textsuperscript{2,3}},
 \textbf{Rattandeep Singh\textsuperscript{1}},
 \textbf{Zhou Yu\textsuperscript{1,3}},
\\
\\
 \textsuperscript{1}Columbia University,
 \textsuperscript{2}Georgia Institute of Technology,
 \textsuperscript{3}Arklex.ai,
\\
 \small{
   \textbf{Correspondence:} \href{mailto:mt3639@columbia.edu}{mt3639@columbia.edu}
 }
}

\begin{document}
\maketitle
\begin{abstract}

End-to-end form filling refers to automatically populating fields in a document-style form with the appropriate information derived from external data.
Although prevalent and useful, no formal benchmark exists for evaluating systems' form completion accuracy.
Existing datasets focus on parsing, extraction and web form interaction, rather than end-to-end completion of document-style forms.
We propose FormGym, a benchmark formulation of the end-to-end form filling task that evaluates form completion and accuracy.
We construct FormGym by repurposing three existing datasets and add one new dataset to achieve more challenging, diverse, and realistic test cases.
Our studies show baseline vision language agents (VLAs) perform poorly on FormGym in every scenario, primarily due to poor field localization.
GUI agents perform better but suffer from high latency and costs.
Therefore we also introduce FieldFinder, a field localization tool that enables zero-shot VLAs to find and accurately place text in input fields.
We find that VLAs augmented with FieldFinder achieve better performance compared to baselines in all models.

\end{abstract}

\section{Introduction}

% Motivation
Filling out paperwork is a pervasive and tedious task.
The US government estimates that federal agencies collectively generate nearly 10 billion hours of mandatory paperwork each year \cite{OMB2016ICB}.
Ad-hoc tools and advances among digital agents in related fields have shown potential to substantially reduce this burden \cite{ghosh2024exploring}.
However, we are unaware of any formal benchmark to evaluate automatic form filling.
This results in uncertainty and risk for users, especially on forms carrying legal or financial weight.
% Our contributions extend existing work in document understanding by addressing the localization of empty input fields in an agentic context.

Existing literature addresses components of end-to-end (E2E) form filling, including document understanding, question answering, tool use, and image manipulation.
Work on GUI agent systems and benchmarks has primarily targeted browser and other desktop UIs.
However, we are concerned with printable, PDF-style images of target documents.
Lacking the same appearance and affordances as GUIs, these documents will challenge existing strategies, especially systems reliant on underlying structures such as the DOM or metadata.

\begin{figure}[t]
    \centering
    \includegraphics[width=1.0\linewidth]{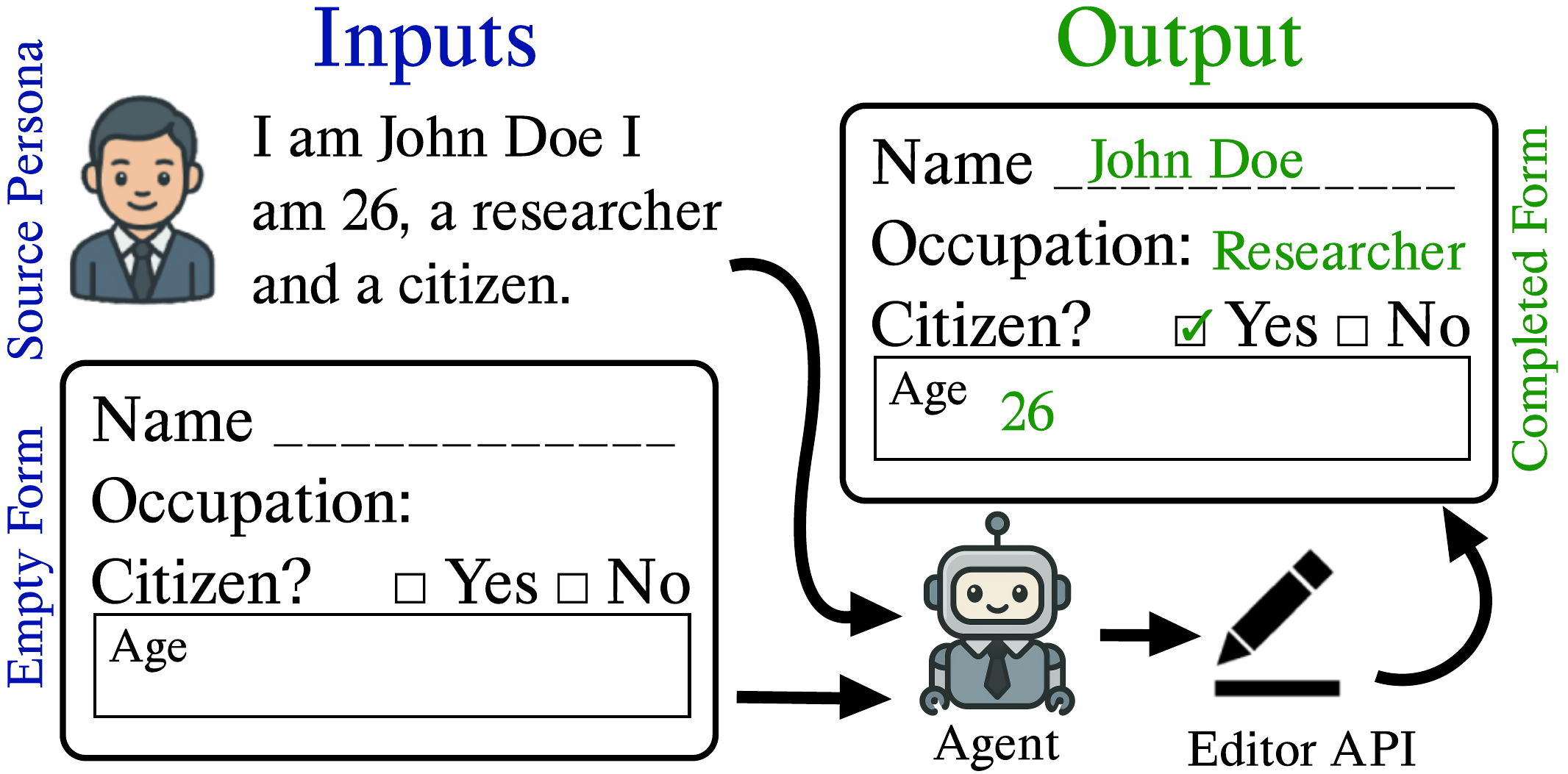}
    \caption{In the FormGym task, agents are provided with an unfilled source form and a user persona containing answers to fields in the form. The agent must use an editor API to produce the completed source form. Diverse layout semantics indicate suggested fields, such as underlines, colons, check boxes, and table cells.}
    \label{fig:task}
\end{figure}

In form filling tasks, each target document contains one or more empty fields, each associated with a field name (e.g. Occupation) in the document indicating the desired field value (e.g. Researcher) through layout semantics (e.g. proximity) (Figure \ref{fig:task}).
Field values are derived from an external source persona, such as free text or an image of a related document.
% We formalize the E2E form filling task as follows: For each field, place a correct string inside its bounding box conditioned on the document, field name, and input data.
We formalize the E2E form filling task mapping unfilled forms and source persona to filled forms as follows: for each field in the unfilled form, place an appropriate string inside its bounding box.

We introduce FormGym, a novel benchmark dataset and evaluation framework for end-to-end form filling.
To construct FormGym, we leverage three existing datasets originally constructed for document understanding.
From these, we constructing ground truth field (name, bounding box, value) tuples for empty forms by redacting values.
% Identifying limited flexibility and complexity in these forms, we manually annotate an additional Auto Loans dataset and construct multiple synthetic user profiles representing potential applicants.
However, because of their original annotation schemes, it is difficult to construct realistic source personas without the field values appearing verbatim.
We therefore extend our benchmark by manually annotating the Auto Loans dataset with a more sophisticated schema.
This allows FormGym to address longer forms, incorporate secondary source documents into source personas, and include fields that require composing values from multiple facts rather than copying text directly.

Noting near-zero accuracy by baseline vision language agents (VLAs), we identify field localization as the main bottleneck in end-to-end form filling, even in frontier models.
Rather than train purpose-built agents, we would prefer to provide flexible infrastructure adaptable to any new LLM.
Hence, we develop FieldFinder, an open vocabulary model trained to output the bounding box of the field associated with an input field name.
As a lightweight tool, FieldFinder allows VLAs to specify the target field by name rather than cartesian coordinates.
We demonstrate that FieldFinder improves VLA form filling accuracy across all scenarios and models with minimal latency and memory overhead (Figure \ref{fig:task}).
Our contributions are twofold:
\begin{itemize}
    \item \noindent\textbf{A benchmark for evaluating agents on end-to-end form completion}, showing that current VLAs and Claude Computer Use struggle to accurately identify field placements.
    \item \noindent\textbf{An open-vocabulary field localization model}, showing that it helps VLAs overcome spatial reasoning limitations.
\end{itemize}

Our code and dataset are available at \href{https://github.com/mtoles/formgym}{https://github.com/mtoles/formgym}

\section{Related Work}

Several benchmarks exist for evaluating document layout \cite{zhong2019publaynet, pfitzmann2022doclaynet, li2020docbank, li2019tablebank, harley2015icdar, li2019tablebank}. 
Numerous frameworks \cite{xu2020layoutlm, li2021selfdoc, bao2020unilmv2, appalaraju2021docformer, lee2022formnet, lu2024omniparser} have been proposed for navigating these types of tasks (Table \ref{tab:related-work}). 
Despite deep exploration of document understanding, prior work has not addressed document elements that are suggested only by negative space in the document (e.g., Occupation in Figure \ref{fig:task}). %, such as the answers to questions and empty table cells.
Some challenging examples include non-underlined fields, table cells or those indicated merely by a colon (e.g., "Name: \quad"). 
% This presents an open question of how to locate these suggested elements, whose boundary at the time of inference may contain no more than an empty white space, and how to evaluate such models.
Existing commercial software, such as Mac OS Preview and Amazon Textract, can localize some but not all suggested elements. %suggested elements such as text fields in image PDFs, but also struggles with these challenging fields. 

% , and graph \citep{yu2021pick, zhang2020trie} transformers
Unlike traditional question answer-style (QA) benchmarks, vision language and graphical user interface (GUI) agent evaluations generally measure a path-independent end-state, such as in \citet{zhou2023webarena}, \citet{zheng2022vlmbench}, \citet{liu2023agentbench}, \citet{yao2024tau}, and \citet{he2024webvoyager}, which often include elements of form-filling.
GUI agents including \cite{qin2025uitarspioneeringautomatedgui, gou2025navigatingdigitalworldhumans} generally focus training on GUIs rather than flat document images and interact with documents through browser UIs rather than editor APIs.
In contrast, our work explores end-to-end, real-world completion of PDF-style image domain forms. %  by focusing on the localization of suggested input text.

Computer use agents including Claude Computer Use \cite{anthropic2024computeruse} and OpenAI Operator  \cite{openai2025operator} show potential to address these types of tasks.
Open source frameworks have also been proposed, including  \citet{Shen2024FalconUI,Qin2025UITARS,Liu2024AutoGLM,Hong2024CogAgent,Wei2025WebAgentR1,Qi2025WebRL,Ma2023LASER,Putta2024AgentQ,Shen2024ScribeAgent}.
However, many of these solutions exploit structured elements such as the DOM, or are limited in their domain or interaction methods.

\section{FormGym Benchmark}

% We derive FormGym document from two sources: existing document relation datasets and specifically sourced auto loans documents.
The FormGym dataset is composed of documents from three existing datasets plus one novel dataset.
While existing datasets represent a large quantity of diverse, multilingual documents, our Auto Loans dataset introduces some of the most challenging scenarios in the form filling domain.

\subsection{Conversion from Existing Datasets}
\label{sec:existing-datasets}

\begin{figure*}[]
    \centering
    \begin{subfigure}[t]{0.24\linewidth}
        \centering
        \includegraphics[width=\linewidth]{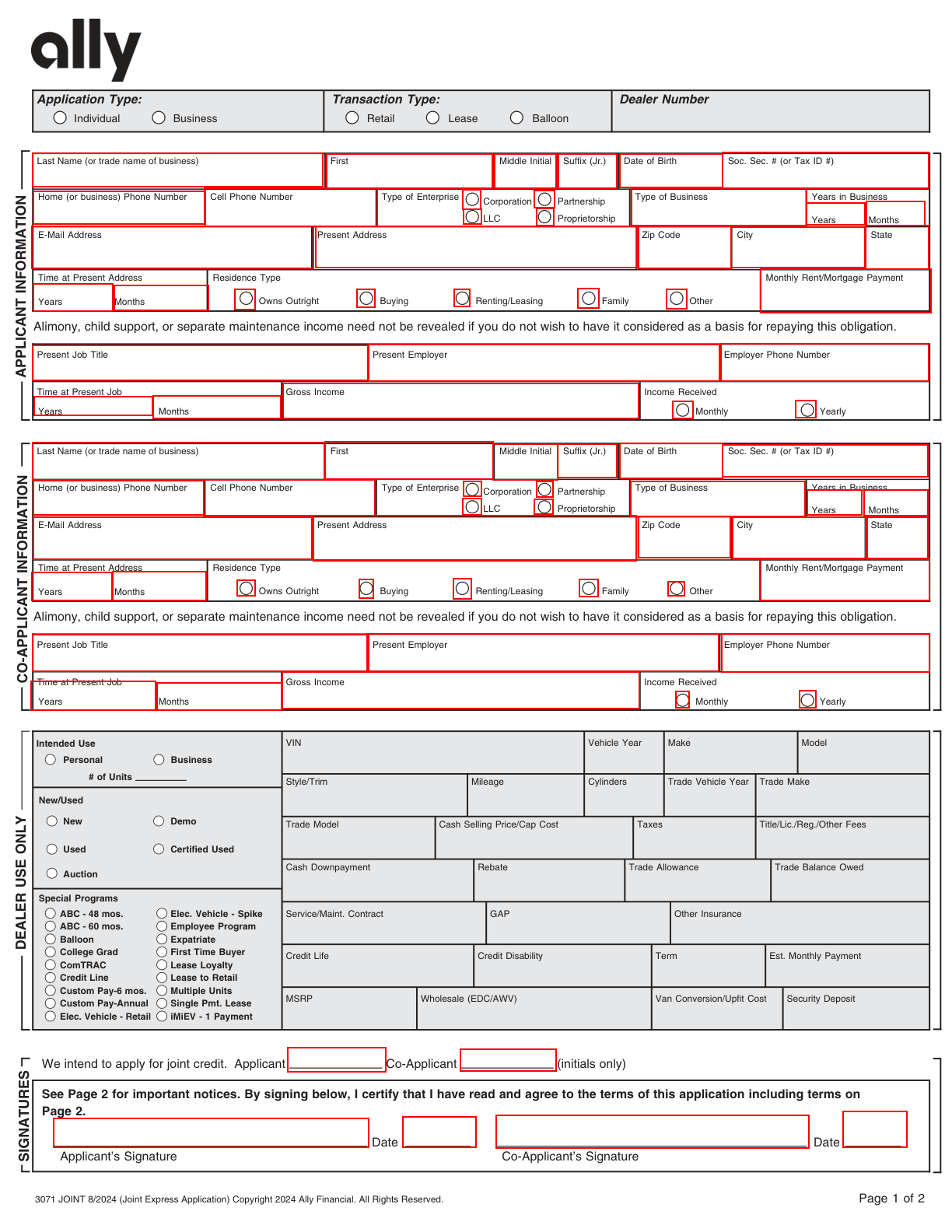}
        \caption{Auto Loans}
        \label{fig:al}
    \end{subfigure}
    \begin{subfigure}[t]{0.24\linewidth}
        \centering
        \includegraphics[width=\linewidth]{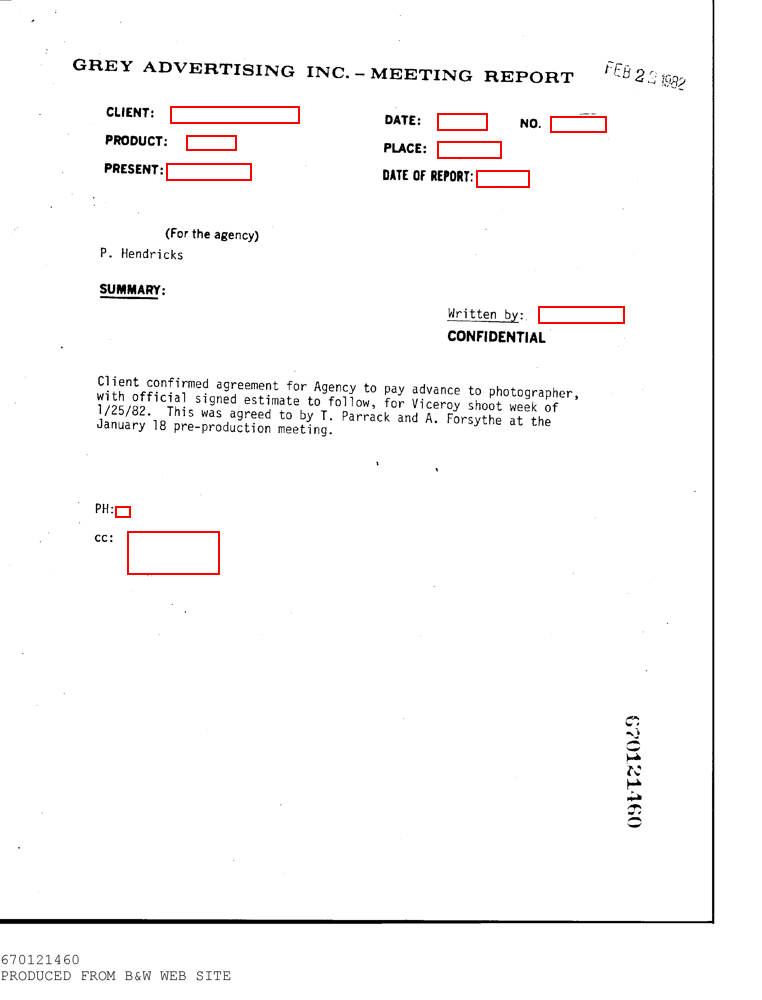}
        \caption{FUNSD}
        \label{fig:funsd}
    \end{subfigure}
    \begin{subfigure}[t]{0.24\linewidth}
        \centering
        \includegraphics[width=\linewidth]{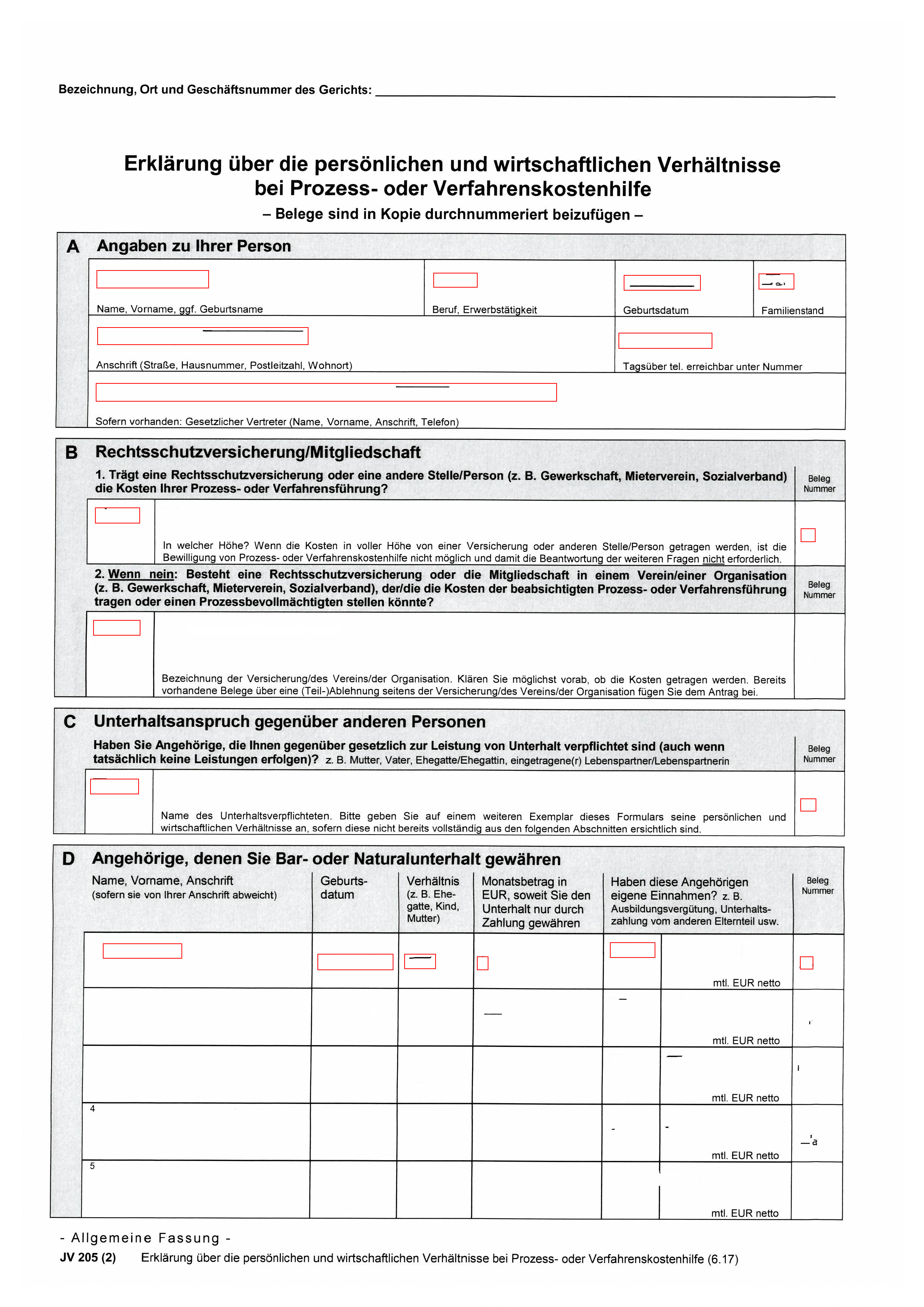}
        \caption{XFUND}
        \label{fig:xfund}
    \end{subfigure}
    \begin{subfigure}[t]{0.24\linewidth}
        \centering
        \includegraphics[width=\linewidth]{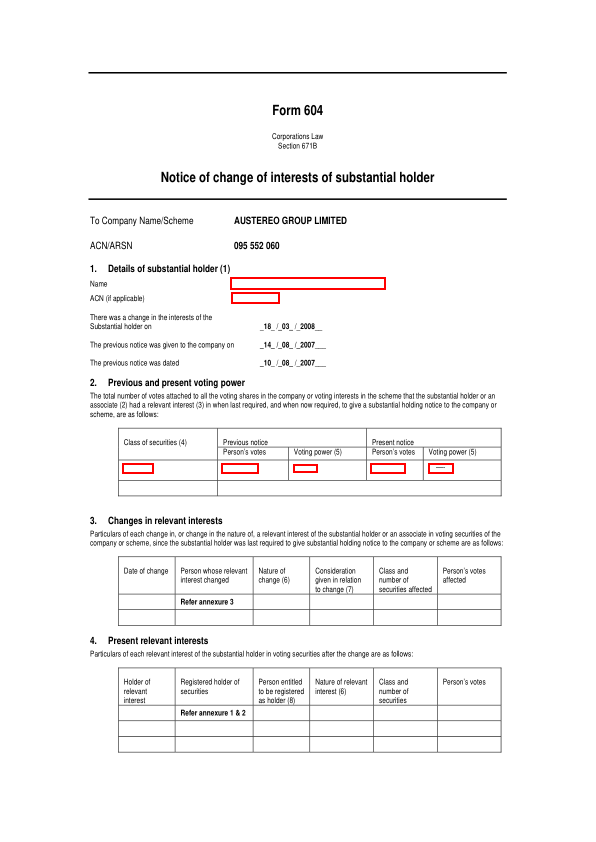}
        \caption{Form-NLU}
        \label{fig:formnlu}
    \end{subfigure}
    \caption{Example forms and field bounding boxes in the FormGym dataset.}
    \label{fig:examples}
\end{figure*}

We draw examples from the FUNSD \cite{jaume2019funsd}, XFUND \cite{xu2022xfund}, and Form-NLU \cite{10.1145/3539618.3591886} document relation datasets, representing scanned English, scanned multilingual, and digital-born English financial documents, respectively (Figure \ref{fig:examples}).
We choose these datasets because they contain a large quantity of relationship annotations between document field names and values, as well their bounding boxes (Table \ref{tab:dataset}).
FUNSD is composed scanned documents from the Truth Tobacco Industry Document archive \cite{UCSF_Tobacco_Industry_Documents}.
XFUND is composed of scanned documents from Common Crawl in seven languages \cite{commoncrawl}.
Form-NLU is composed of Australian financial filings.
For each key-value relation, we create one (document image, field name, field value, value bounding box) example.
We extract names, values, and bounding boxes from document annotations.
On these documents, the correctness of predicted field values is determined by exact match.
Because these datasets include only completed documents, we then create empty fillable documents by deleting value text from filled fields with horizontal inward content-aware fill.\footnote{github.com/light-and-ray/resynthesizer-python-lib}

%If the table is too wide, replace \begin{table}[!htp]...\end{table} with
%\begin{adjustwidth}{-2.5 cm}{-2.5 cm}\centering\begin{threeparttable}[!htb]...\end{threeparttable}\end{adjustwidth}
\begin{table*}[!h]\centering
\small
\setlength{\tabcolsep}{4pt} % default ~6pt
\begin{tabular}{lrrrrrrrrrrrr}\toprule
&\multicolumn{2}{c}{\textbf{Forms}} &\multicolumn{2}{c}{\textbf{Fields}} &\multicolumn{2}{c}{\textbf{Fields/Form}} &\textbf{Sources} &\textbf{Lang.} &\textbf{Domain} &\textbf{Quality} &\textbf{Comp.} \\\midrule
&\textbf{Train} &\textbf{Test} &\textbf{Train} &\textbf{Test} &\textbf{Train} &\textbf{Test} &\textbf{} & & & & \\
\midrule
Auto Loans &- &10 &- &886 & &88.6 &4 &1 &Auto Loans &Digital &\checkmark \\
FUNSD &155 &39 &2,246 &577 &14.5 &14.8 &1 &1 &US Government &Scanned & \\
XFUND &1,112 &100 &19,559 &1,950 &17.6 &19.5 &1 &7 &Common Crawl &Scanned & \\
Form-NLU &442 &66 &3,661 &476 &8.3 &7.2 &1 &1 &AU Stock Exchange &Digital & \\
\midrule
Total &1,709 &215 &25,466 &3,889 &14.9 &18.1 & & & & & \\
\bottomrule
\end{tabular}
\caption{FormGym Dataset statistics. Sources indicates the number of source personas for field value and prompt generation. Lang. indicates the number of languages. Comp. indicates whether field values must be generated by composing from multiple facts rather than merely copying a span.}
\label{tab:dataset}
\end{table*}

\subsection{Creating A New Auto Loans Dataset}
\label{sec:auto-loans}

While the above documents present a diversity of forms, the automatic nature of example creation does not explore the full complexity of real-world form filling.
Specifically, field names and values may be phrased differently from the source persona (e.g. "previous address" vs. "last place of residence") and may require generation based on multiple facts, such as a street, city, and postal code.
In addition, forms in existing datasets contain relatively few fields per form, substantially reducing the challenge of field localization.
% Source information may not be readily available in the plain text context and instead could come from another document.
Finally, users may wish to have forms completed based on the content of other document images to avoid transcribing relevant information. 
In this case, we will require multiple forms containing partially overlapping field values.
Therefore, we construct the additional Auto Loans dataset.

% \subsubsection{Auto Loans Dataset Collection and Annotation}

We searched the web for American auto loan application PDF files, choosing the single page with the most fields.
One author then manually annotated the bounding box of each empty field.
Another author reviewed these annotations and together they discussed and corrected any disagreements.
We present additional annotation details in (Appendix \ref{app:annotation}).
We defined multiple source personas ensuring that all information necessary for every form be present, such as the applicant's first, middle, and last name.
This increases the diversity of source personas and enables better coverage of certain form fields, including check boxes, which are are sparsely filled.
Observing that most enumerated options in these fields have four or fewer choices, we construct four personas, each spanning all documents.
Finally, for each field, we define a function that determines prediction correctness based on one or more source persona elements.
For example, the Full Name field will be marked as correct iff \texttt{input == source\_data.firstname + " " + source\_data.lastname}.

% \subsubsection{Auto Loans Additional Features}

Defining source personas independently from form fields allows several forms of flexibility not available in XFUND, FUNSD, and Form-NLU.
Specifically, it allows us to accept a broader range of inputs, for example, different phone number formats.
It also permits automatically propagating other documents to serve as ground truth source persona inputs for the Auto Loans (Image) task (Section \ref{sec:auto-loans}).
By manually annotating empty forms instead of automatically redacting completed ones, we avoid artifacts from content-aware fill and biases from original text placement. 
This yields a subset with maximal evaluation precision.

\subsection{FormGym Dataset Statistics}
In total, FormGym contains 25,466 train and 3,889 test field examples spread across 1,709 train and 215 test forms (Table \ref{tab:dataset}).
The dataset contains between 7.2 and 88.6 fields per form with an average of 18.1.
46\% of test documents are in one of seven non-English languages.
65\% of documents are scanned versus digital-born.

\section{FieldFinder Field Localizer}
\begin{figure}[t]
    \centering
    \includegraphics[width=1.0\linewidth]{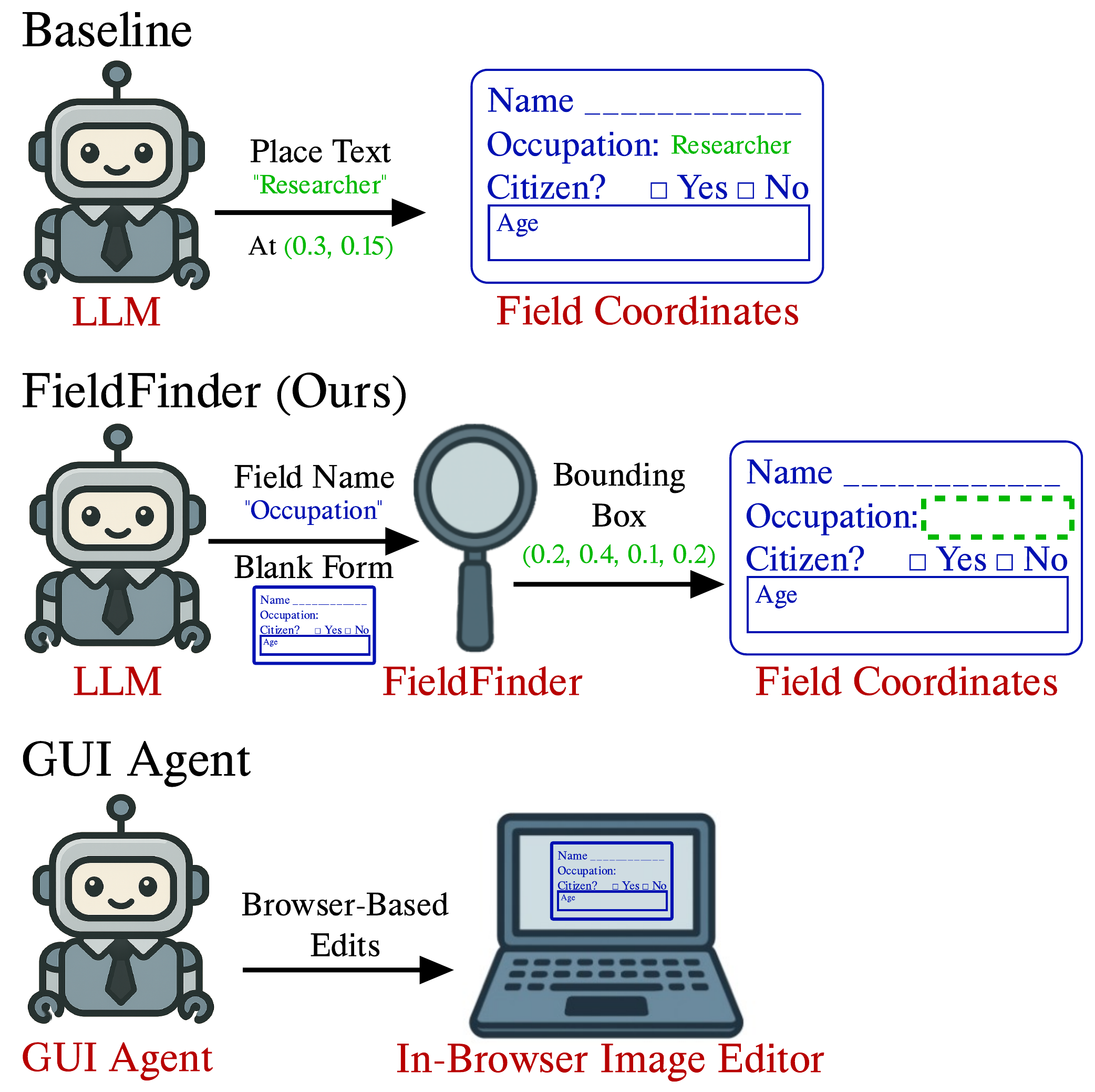}
    % \caption{The FieldFinder tool receives an unfilled source form and the name of a target field ("Occupation") as \textcolor[HTML]{0011b2}{inputs} from an agent. It must \textcolor[HTML]{00b900}{output} the bounding box (dashed box) around the field corresponding to the target field name.}
    \caption{In the baseline case, the LLM receives an unfilled source form and persona information in its context and attempts to complete the form through a text placement API based on (x, y) coordinates. In the FieldFinder (ours) case, the text placement API is replaced with the FieldFinder tool. Instead of using coordinates as input, the FieldFinder tool takes the name of a field as input, then uses an open vocabulary object detection model to detect the corresponding field boudning box. In the GUI Agent case, the GUI agent uses an in-browser image editing tool (designed for humans) to place text on the PDf.}
    \label{fig:tool}
\end{figure}
We find that baseline VLAs perform poorly, primarily due to poor field localization.
We therefore design a field localization tool called FieldFinder that augments any VLA by taking on the localization task.
FieldFinder is trained to take a form image and text description of the name of the target field as input and predicts the bounding box around the valid input space (Figure \ref{fig:tool}). 
Functionally, FieldFinder is an open-vocabulary object detection model, where the target objects are suggested text input fields associated with nearby text that matches the input string.
% We allow VLAs to use FieldFinder in the API by replacing \texttt{PlaceText} with the following, and provide a similar endpoint to replace \texttt{SignOrInitial}:\\

% \noindent\textbf{\texttt{FieldFinder(image, field\_name, text)}} - Place the text \texttt{text} in the field corresponding to \texttt{field\_name} on \texttt{image}.\\

\subsection{FieldFinder Dataset}
\label{sec:methods.dataset}
We draw examples from the FUNSD, XFUND, and Form-NLU train datasets (Figure \ref{fig:examples}).
In some cases, field names can only be described precisely by including additional hierarchical information, such as section headers.
For example, in several Auto Loans forms, personal information for the applicant and the coapplicant are only indicated based on a distant section header.
Similarly, specific table cells must be described by row, column, and table headers.
We therefore prepend descriptions of all hierarchically superior elements to the input text (i.e., \texttt{Section 1 | Members Table | Names Column | Row 1 $\rightarrow$ John Doe }).
Examples consist of (field name, field bounding box) pairs such that FieldFinder can learn to localize the bounding box of a field's value given its name.

\subsection{FieldFinder Training}

We fine-tune a Florence 2 Large \cite{xiao2024florence} vision foundation model to predict the field bounding box given the target field name string and form image.
We choose Florence 2 because its pretraining contains both open-vocabulary object detection and tasks requiring OCR, minimizing the distribution shift between pretraining and fine-tuning.
Florence 2 Large has only 0.77B parameters, contributing minimal latency and memory overhead when augmenting with much larger VLAs.
We train the FieldFinder for 6 epochs, batch size 8, learning rate 1e-6 with cosine profile on 1x NVIDIA A100 GPU for 30 hours.
We selected these parameters using grid search across 2-4 options per parameter.
% The FieldFinder achieves an intersect-over-union of 19.9\% on the FUNSD test set.

\subsection{FieldFinder Results}
\label{sec:ff-results}

\begin{figure}[t]
    \centering
    \includegraphics[width=1.0\linewidth]{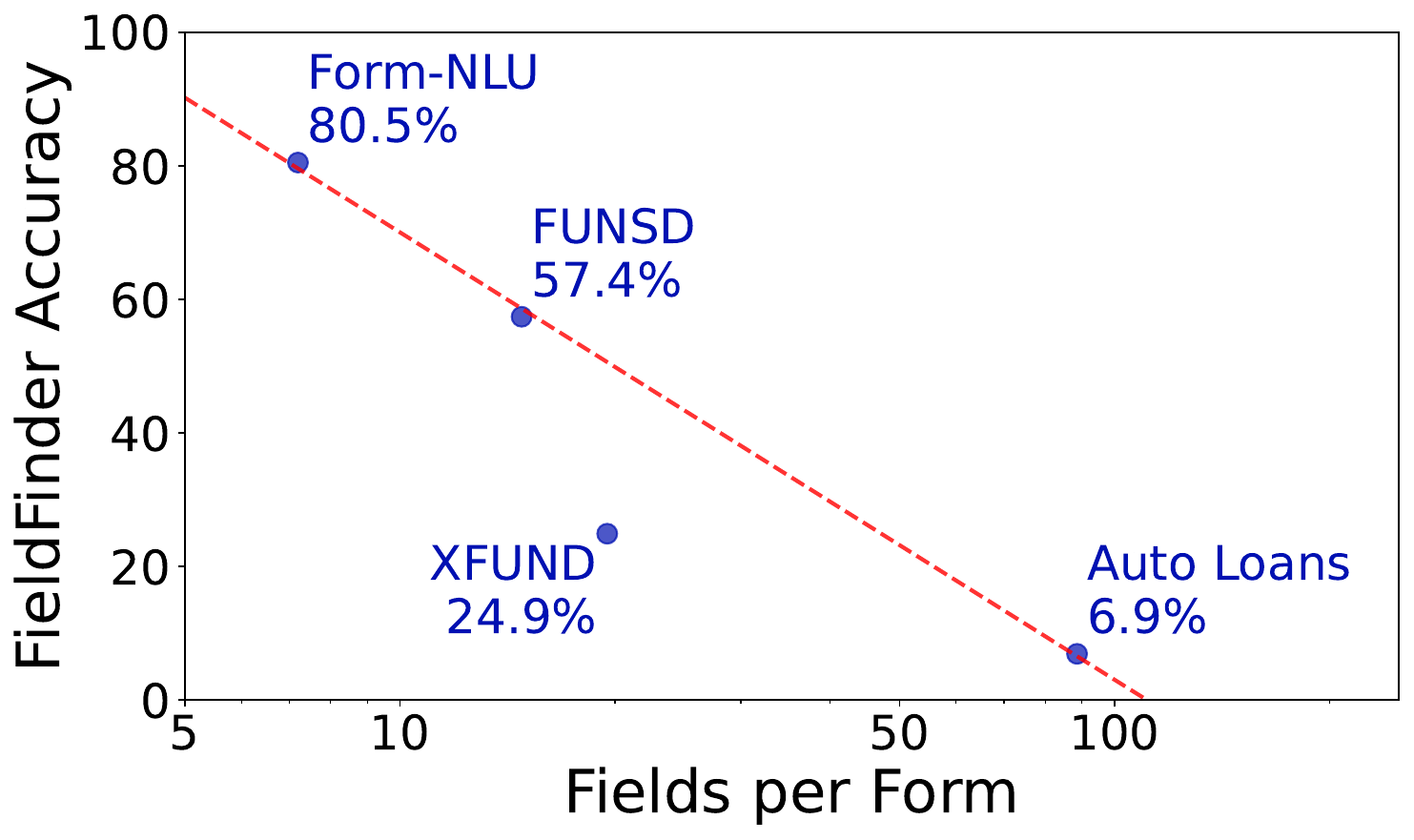}
    \caption{FieldFinder accuracy vs. fields per form (log scale). Trend line shown for English datasets only. The trend suggests that high numbers of fields per form and multi-lingual forms are the greatest challenges for FieldFinder.}
    \label{fig:acc_vs_fpf}
\end{figure}

We choose accuracy as our primary metric, defined as whether the center of the predicted bounding box falls within the ground truth bounding box.
FieldFinder achives 54.3\% accuracy on average across the the test sets of all FUNSD, XFUND, and Form-NLU, compared to near-zero accuracy by baseline VLAs.
% The center of the FieldFiender prediction is within the target bounding box of training distribution holdout cases 42.4\% of the time.
However, we note a wide disparity in performance depending on the dataset (Figure \ref{fig:acc_vs_fpf}).
% To study the influence of FieldFinder errors on overall accuracy, we analyze FieldFinder accuracy on the four datasets given ground truth form text rather than VLA tool calls, which may be unfaithful to the document and context. 
We see the strongest performance in Form-NLU at 80.5\% accuracy, which we attribute to its few fields per form and lack of scan artifacts.
Accuracy decreases on FUNSD to 57.4\%, which contain low quality scans.
XFUND presents an even greater challenge due to its multilingual scans, with FieldFinder reaching 24.9\% accuracy. 
Finally, we find weak performance on Auto Loans at only 6.9\% accuracy, which did not appear in the training data and contains by far the most fields per form, at 88.6 fields per form.
We conclude that FieldFinder can perform well on high quality English documents with few fields per form, but struggles in other contexts and out-of-distribution data.

% However, we cannot confidently attribute these errors to distribution shift from training data containing fewer fields per page.
Although distribution shift likely affects quality on Auto Loans, we draw attention to a confounding factor;
Auto Loans documents are so densely covered with fillable fields (88.6 per page) that there is a high probability for an errant placement to land in an unrelated field, resulting in two errors---an empty filed and a doubly filled field.
When accuracy is plotted against fields per form on a log scale, English datasets Form-NLU, FUNSD, and Auto Loans shows a clear negative linear trend (Figure \ref{fig:acc_vs_fpf}).
This further indicates that FieldFinder accuracy highly dependent on the number of fields per page.
% A partial explanation is that FieldFinder can accurately identify field candidates, but struggles to differentiate which correspond to each name.

\section{End-to-End Form Filling Pipelines}

Noting the strong performance of VLAs on related visual reasoning and QA tasks, we instantiate an E2E form filling VLA pipeline for their evaluation.
While many methods have been proposed for related tasks, only VLAs possess the necessary capabilities to fully perform end-to-end form filling (Appendix \ref{app:existing-methods-comparison}).
We construct parallel pipelines for VLAs augmented with our FieldFinder tool and Claude Comptuer Use GUI agent.

\subsection{VLA Pipeline}
\label{sec:vla-pipeline}
% \subsubsection{Baseline VLAs}

We choose leading open- and closed-source VLAs as our baselines because they represent the only end-to-end solutions that the necessary criteria in table \ref{tab:related-work}.
We select Llava 7B, Molmo 7B, and Aria 25B as our open-source; and Claude 4 and GPT-5 as our frontier VLAs.
We prompt VLAs with the source image, source persona and form editing API documentation.
We also evaluate Claude 4 + Set of Marks \cite{yang2023setofmarkpromptingunleashesextraordinary}, overlaying a grid with coordinate references at vertices.

% \subsubsection{Form Editing API}
To leverage VLAs existing tool use capabilities, we construct a minimal image editing API by which they can edit the form. 
The image editing API permits placing text or signatures/initials on the image by specifying the string and (x, y) coordinates.
It also allows for deleting all input text intersecting an (x, y) coordinate and terminating a session early.

% \noindent\textbf{\texttt{PlaceText(image, x, y, text)}} - Place the text \texttt{text} on \texttt{image} centered at the coordinates \((x, y)\).
  
% \noindent\textbf{\texttt{SignOrInitial(image, x, y, text)}} - Place the \texttt{text} at coordinate \((x, y)\) on \texttt{image} in the form of a signature or initials.
 
% \noindent\textbf{\texttt{DeleteText(image, x, y)}} - Delete all input text on \texttt{image} whose bounding boxes contain the coordinate \((x, y)\).
 
% \noindent\textbf{\texttt{Terminate()}} - End the current session.

\subsection{FieldFinder Pipeline}

We evaluate VLAs augmented with FieldFinder in a pipeline consistent with the previously described VLA pipeline.
However, we replace the text and signature placement APIs with a FieldFinder API.
Rather than taking Cartesian coordinates as input along with the field value string, the FieldFinder API takes a field name and value. 
It then places the value at the location detected by FieldFinder based on the field name.

\subsection{GUI Agent Pipeline}
We instantiate the Claude Computer use Use GUI agent with the free in-browser photo editing application Photopea\footnote{photopea.com}, whose interface is nearly identical to Adobe Photoshop.
We prompt Claude Computer Use with the natural language source persona and instructions to complete the form.
Prompts include detailed instructions on how to use the Photopea interface, without which GUI agents fail completely.
We provide additional implementation details in Appendix \ref{sec:example_gui_prompt}.
For accessibility and cost reasons, we limit operators to five minutes per page and downsample test sets to 10 documents per source.

\begin{table*}[]\centering
\small
\begin{tabular}{lrrrrrrrrrr}\toprule
&\multicolumn{2}{c}{\textbf{AL (Text)}} &\multicolumn{2}{c}{\textbf{AL (Image)}} &\textbf{FUNSD} &\textbf{XFUND} &\textbf{Form-NLU} &\textbf{Average} &\textbf{Cost} \\\cmidrule{2-10}
&\textbf{OS} &\textbf{IT} &\textbf{OS} &\textbf{IT} &\textbf{OS} &\textbf{OS} &\textbf{OS} & & \\ \midrule
Aria 25B &0.0 &0.0 &0.0 &0.0 &1.0 &1.0 &0.0 &0.4 &- \\
Claude 4 &0.0 &0.0 &0.3 &0.7 &21.0 &1.0 &0.0 &4.5 &0.76 \\
GPT-5 &1.0 &1.5 &0.5 &0.7 &2.0 &2.0 &0.0 &1.2 &\textbf{0.30} \\
Llava 7B &0.0 &0.0 &0.0 &0.0 &1.0 &1.0 &0.0 &0.4 &- \\
Molmo 7B &0.0 &0.0 &- &- &1.0 &1.0 &0.0 &0.5 &- \\
Claude 4 + SoM &0.0 &1.0 &1.0 &1.0 &9.0 &1.0 &0.0 &2.3 &0.69 \\
\midrule
Aria 25B + FF (ours) &3.3 &4.2 &1.5 &2.0 &20.0 &9.0 &29.0 &12.7 &- \\
Claude 4 + FF (ours) &8.3 &9.2 &\textbf{4.8} &\textbf{5.3} &32.0 &15.0 &\textbf{54.0} &\textbf{23.0} &0.43 \\
GPT-5 + FF (ours) &\textbf{8.5} &\textbf{9.8} &3.0 &\textbf{5.3} &29.0 &14.0 &50.0 &21.3 &0.37 \\
Llava 7B + FF (ours) &1.0 &1.0 &0.3 &0.3 &4.0 &3.0 &4.0 &2.5 &- \\
Molmo 7B + FF (ours) &0.5 &1.3 &- &- &9.0 &3.0 &7.0 &5.0 &- \\
\midrule
Claude Computer Use &2.7 &- &1.4 &- &\textbf{46.2} &\textbf{26.7} &32.4 &21.1 &54.00 \\
\midrule
Claude 4 (GT Coords) &74.0 &90.0 &49.2 &56.2 &83.0 &82.0 &77.0 &75.3 &- \\
\bottomrule
\end{tabular}
\caption{Percentage of fields containing a correct value. AL: Auto Loans dataset. IT: Iterative. OS: One-Shot. FF: FieldFinder. SoM: Set of Marks. GT: VLA was supplied with ground truth field coordinates in the prompt. Macro average is taken across each dataset. Cost indicates API fees in USD per thousand fields. We do not include Molmo in the Auto Loans (Image) task because it is not pretrained with multiple images.}
\label{tab:results}
\end{table*}
\section{Experiments and Results}
\label{sec:results}

\subsection{Pipeline Settings}
\label{sec:pipeline-settings}

Because Auto Loans documents contain numerous fields per page, we evaluate VLAs under two settings:

\noindent\textbf{One-Shot} - The agent must place all text at once.

\noindent\textbf{Iterative} - The agent may take multiple sets of actions over the course of up to 5 rounds, allowing it to correct mistakes. 
Agents receive feedback on the success or failure of each action.

The Iterative setting allows agents to adapt to tooling and to realize and correct mistakes using the deletion API endpoint.
The GUI agent pipeline is inherently iterative.
We present source personas in two forms:

\noindent\textbf{Text} - The source persona is presented as plain text.

\noindent\textbf{Image} - The source persona is presented as an image of another auto loan document. Persona data not appearing in the image is presented in plain text.

We evaluate VLAs in our E2E form filling pipeline with and without FieldFinder on the FormGym dataset.
We find FieldFinder improves accuracy in all cases.

\subsection{Metrics}

We evaluate field completion correctness based on whether each field contains the correct value, as defined in Sections \ref{sec:existing-datasets} and \ref{sec:auto-loans}. 
If a field contains multiple text inputs, we concatenate them with a space.
We choose field accuracy as our primary evaluation metric, ignoring those that should be empty according to the ground truth label to avoid inflating accuracy. 
A text input is considered to be inside a field if the center point of the text bounding box is within the field.
A side effect of this rule is the increased difficulty of forms with numerous fields.
On such forms, inaccurate placements are more likely to fall into unrelated fields, resulting in a double error.

We choose center point correctness instead of other metrics including intersect-over-union or full enclosure because it does not rely on \textit{a priori} knowledge of text size for models to accurately place text. 
It also avoids ambiguity in how to judge text that slightly overhangs out of a field or into another field when, to a human, the intention of the form filler would be obvious.

\subsection{Performance Comparison}

Overall, VLAs struggle on FormGym, with models performing best on Form-NLU and worst on Auto Loans (Table \ref{tab:results}). 
Baseline VLAs generally score $\leq$ 3\%.
Set of Marks VLA augmentation offers no improvement over baseline VLA setup.
Claude Comptuer Use, however, achieves an average accuracy of 21.1\%
Our method using Claude 4 + FieldFinder, achieves the highest accuracy at 23.0\%.

% In some cases, Claude appears capable of performing vertical but not horizontal localization, frequently generating $x=0.5$. 

One outlier is baseline VLA Claude 4's performance on FUNSD, achieving 21.0\%.
It appears that, in some cases, Claude can performing vertical but not horizontal localization, frequently generating $x=0.5$. 
Because FUNSD fields are very wide, this results in higher accuracy.
We observe a similar effect in Claude Computer Use, which achieves 60.2\% accuracy.
However, when introducing FieldFinder, we observe equal or better performance in all cases (Figure \ref{tab:results}).
% Models with FieldFinder demonstrate particularly strong improvement on Form-NLU, reaching 54
\begin{figure}[t]
    % \centering
    \begin{subfigure}{\linewidth}
        \caption{Baseline}
        \includegraphics[width=\linewidth]{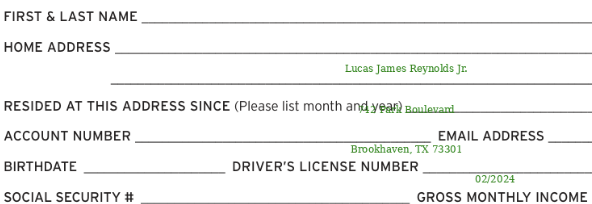}
        % \centering
    \end{subfigure}\\[1ex]
    \begin{subfigure}{\linewidth}
        \caption{With FieldFinder}
        \includegraphics[width=\linewidth]{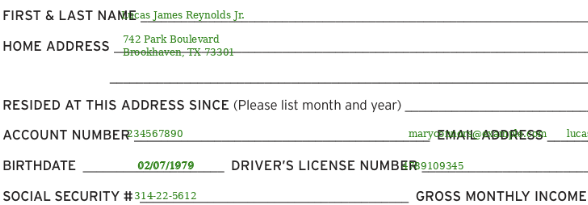}
        % \centering
    \end{subfigure}\\[1ex]
    \begin{subfigure}{\linewidth}
        \caption{Ground Truth}
        \includegraphics[width=\linewidth]{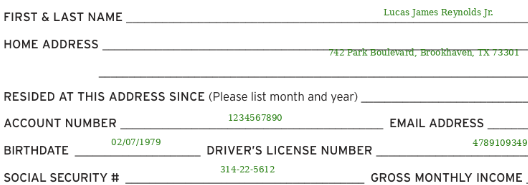}
        % \centering
    \end{subfigure}
    \caption{Example output by Claude 4 baseline, with FieldFinder (ours), and ground truth in the Auto Loans (Text) One-Shot task. We attribute FieldFinder’s leftward bias to supervision artifacts: training labels mark left-biased value text rather than full fields. Without FieldFinder, Claude appears to struggle more with horizontal spacing that with vertical spacing, assigning most placements an x coordinate of exactly 0.5 (not centered due to figure cropping).}
    \label{fig:stacked_images}
\end{figure}
Augmenting with FieldFinder, Claude 4 improves from 4.5 to 23.0\% average accuracy.
In the best case, Claude 4's performance on Form-NLU increases from 0\% to 54\%.
We observe smaller gains, up to 9.2\% percentage points on Auto Loans (Text) (Figure \ref{fig:stacked_images}).
Notably, Claude 4 with FieldFinder surpasses Claude Computer Use on both digital-born datasets, Auto Loans and Form-NLU.
All open-source models also achieve performance improvements with FieldFinder, with Aria leveraging the tool better than smaller models.
GPT-5 and Claude appear to struggle to chained reasoning in the more complex Auto Loans (Image) task, where accuracy falls by around 50\% compared to when source information is supplied in text form.

\begin{figure}[t]
    \centering
    \includegraphics[width=1.0\linewidth]{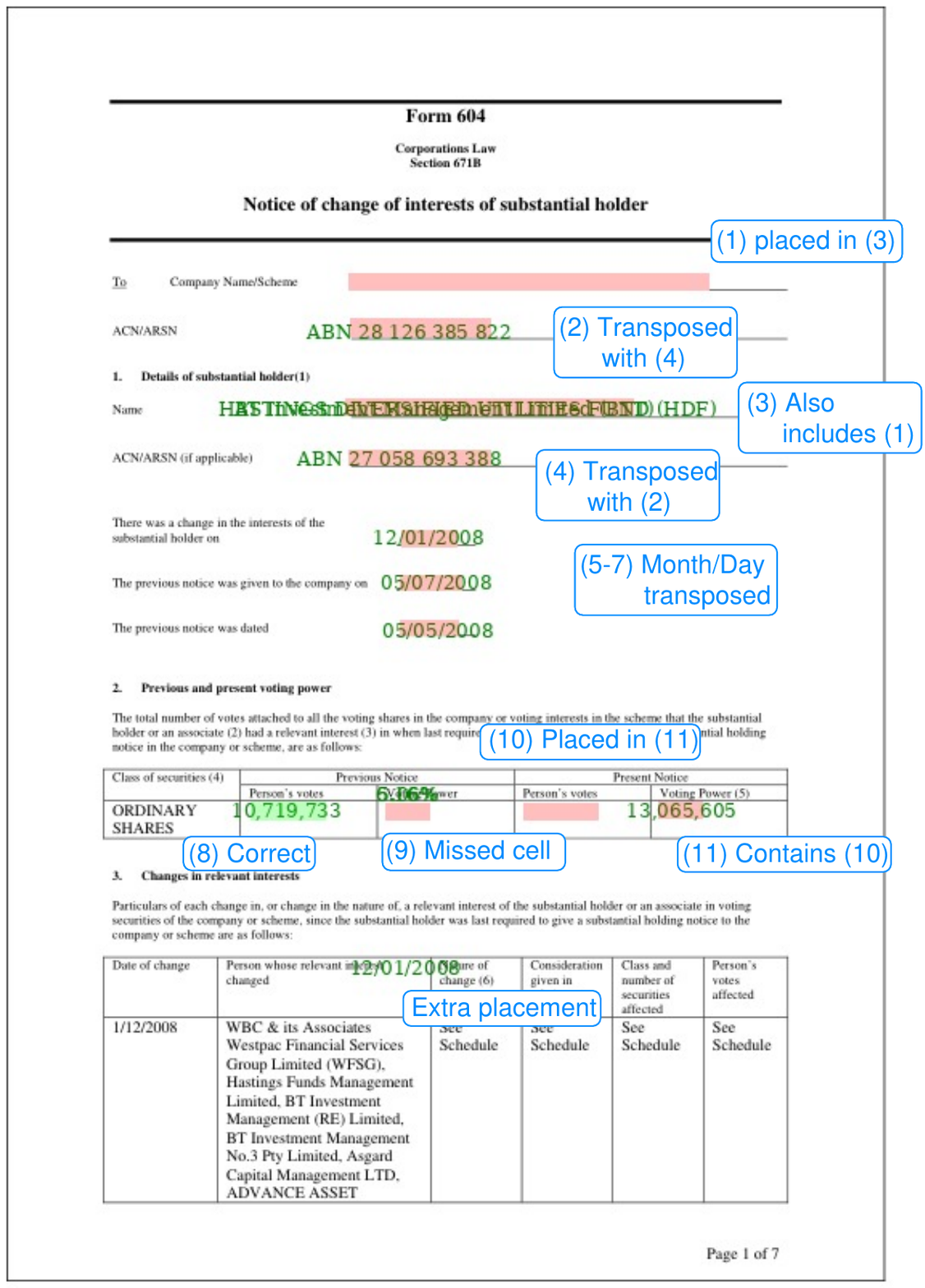}
    \caption{Example failure modes using FieldFinder in an example from Form-NLU. The answer for Field 1 was inaccurately superimposed on Field 3 which is otherwise correct. Answers for Fields 2 and 4 were transposed. Answers in fields 5-7 used an incorrect date format. Field 8 is correct. The answer for Field 9 is outside the field. The answer for Field 10 is placed in Field 11.}
    \label{fig:failure_modes}
\end{figure}

We illustrate the most common FieldFinder errors in Figure \ref{fig:failure_modes}. Primary failure modes include placing text slightly outside the field and placing text in the wrong field with a lexically similar name. The later case can generate secondary failure by adding additional text into the incorrect field, sometimes superimposing it on the otherwise correct answer. 

Our FieldFinder localization tool demonstrates a method for equipping zero-shot VLAs with field localization abilities.
Without FieldFinder, VLAs struggle to even approach this task due to a lack of training data involving pixel-level predictions, with frontier models scoring below 1\% accuracy in multiple domains.
As expected, FieldFinder performs strongest on high quality English document images with relatively few fields.
FieldFinder uses only 0.77B parameters, requiring negligible memory and latency overhead when augmenting both closed- and open-source models with no need for additional training.

We note that XFUND accuracy falls substantially below the English dataset trend line, despite constituting the largest portion of training data.
We attribute this to the multi-lingual nature of the XFUND dataset, a well studied weakness in models pre-trained primarily on English data \cite{geigle-etal-2025-centurio}.
\section{Error Analysis}

Surprisingly, we see little improvement between One-Shot and Iterative flows, except in the case of Aria, which shows a 1.27-fold increase in accuracy (Table \ref{tab:results}).
Analyzing VLA trajectories, we attribute this to frontier models calling the \texttt{Terminate} action after the first turn in most (or in the case of Claude, all) trajectories.
Although these models are prompted with the number of remaining opportunities to edit the form, they appear to lack the self-awareness to doubt their own accuracy and utilize future turns for error correction.

GUI agents present one path to end-to-end form filling, with Claude Computer Use achieving 21.1\% accuracy.
% Claude Computer Use achieves substantially worse performance on the Auto Loans dataset at only 2.7\% accuracy.
Qualitatively, we observe that it is particularly challenged by long documents, often failing to finish within the allotted time. 
It also typically fails on forms containing tables, resulting in only 2.7\% accuracy on Auto Loans (Text).
Overall, Claude Computer Use falls short of our Claude 4 + FieldFinder method (23.0\%).
Moreover, Claude Computer Use costs \$5.40 per 1000 fields, compared to \$0.043 for Claude 4 + FF.
Furthermore, Claude Computer Use requires 28 minutes per 100 fields, compared to approximately 20 seconds seconds for Claude 4.
Ultimately, we find that Claude Computer Use requires over 100x the cost and 84x the time to complete forms while providing worse overall accuracy.

\begin{figure*}[]
    \centering
    \includegraphics[width=0.8\linewidth]{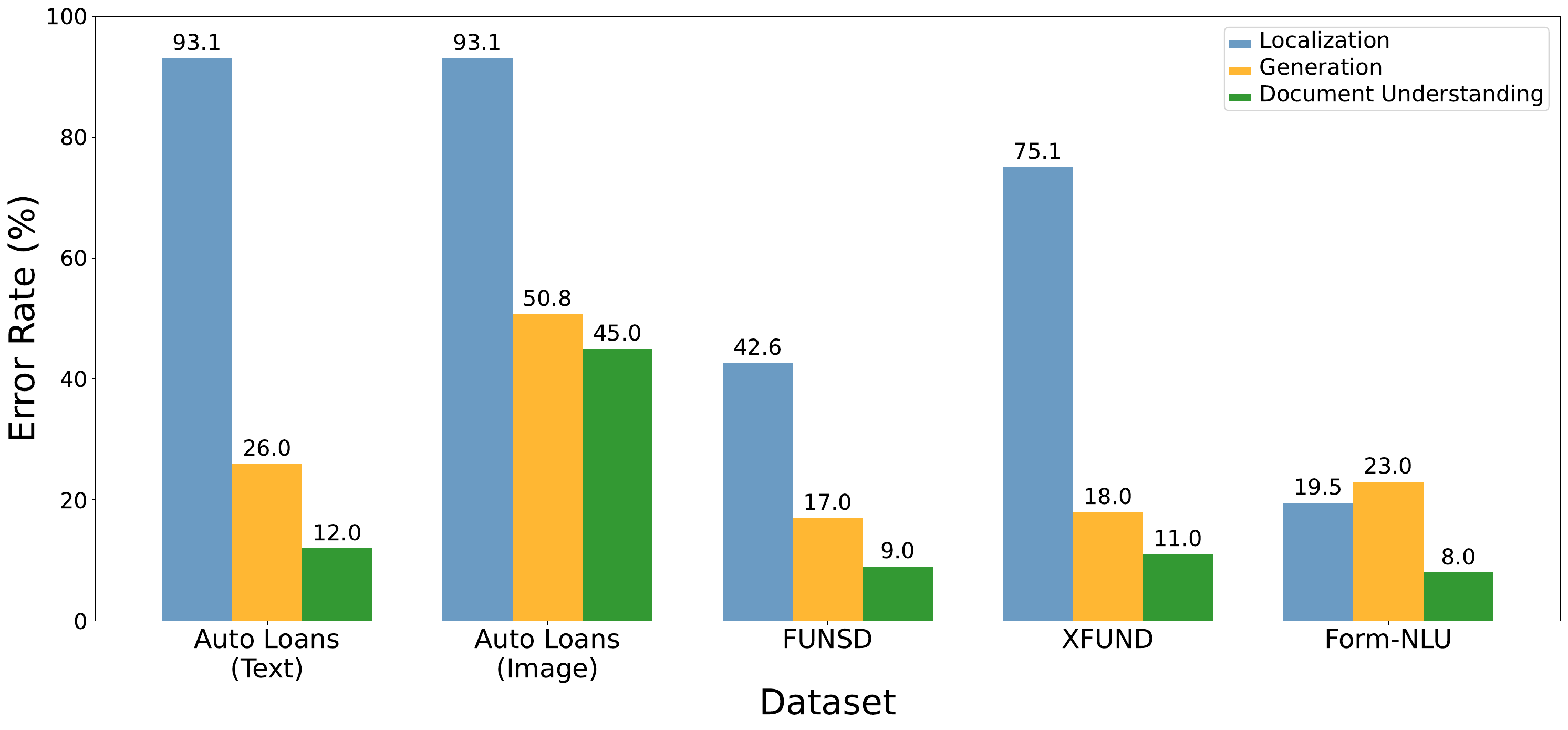}
    \caption{
        Error rates from generation (VLA performance given ground truth coordinates), document understanding (difference between number of placements and fields), and localization (FieldFinder accuracy given ground truth field names). Localization error shown applies only to our FieldFinder pipeline, whereas generation and document understanding errors apply to both FieldFinder and baseline VLA pipelines instantiated with Claude 4.
        }
    \label{fig:ff_vs_gt_coords}
\end{figure*}

% We conduct studies to attribute errors to either the localization (FieldFinder) or document reasoning (VLA) component of the form filling agent.
We conduct studes to attribute errors to either the document understanding or document reasoning steps of the VLA pipeline using Claude 4.
Document understanding errors arise from VLAs failing to understand the input document image.
Document reasoning errors, in contrast, arise from erroneous generations by the VLA even when provided with ground truth free text descriptions of field names and locations.
Compared to the localization errors discussed in Section \ref{sec:ff-results}, these errors are somewhat less common (Figure \ref{fig:ff_vs_gt_coords}) overall, but more common in the high quality, low field density Form-NLU dataset.
\subsection{Document Reasoning Error}

To study the influence of VLA document reasoning errors, we provide the overall strongest model, Claude 4, with ground truth coordinates to the centroid of each field and its name.
Under these conditions, the VLA no longer needs to localize suggested fields, must only map information in the context to the appropriate bounding box.
In general, accuracy is relatively good, ranging from 74\% Auto Loans (Text) to 83\% (FUNSD).
However, accuracy drops from 74\% to 49\% when the source persona is provided in image form (Figure \ref{fig:ff_vs_gt_coords}).
This indicates that, although Claude can understand documents well in simple tasks, it struggles to apply these abilities when they are integrated into a more complex workflow.

\subsection{Document Understanding Error}

In a third failure mode, models may fail to detect fields entirely, resulting in neither a localization nor a reasoning error.
In Figure \ref{fig:ff_vs_gt_coords}, we plot document understanding errors as the absolute difference between the number of text placements and the number of fields.
We observe that Claude and GPT-5 generate as few as 0.42 placements per field in the Auto Loans task, indicating they did not even attempt to fill more than half of fields. 
This tendency is compounded by a frequent early termination rate in Iterative tasks, resulting in a relatively modest improvement between One-Shot and Iterative tasks.
Furthermore, despite being equipped with the delete action, we find that its use is vanishingly rare.

\subsection{Error Comparison}

We observe that even with FieldFinder, localization remains the largest source of errors in all datasets except for Form-NLU.
We note that these failure modes are not well-correlated, suggesting that future work may require a multi-faceted approach.
Additionally, language does not appear to affect document understanding or reasoning significantly, with comparable results across XFUND and FUNSD.
Therefore, we encourage future work addressing document understanding, reasoning, and localization in VLAs.

% Our ablations identify two types of errors occurring in the FieldFinder workflow: localization errors (when text is placed at the wrong location) and document reasoning errors (those attributable to the backbone model text generation, including document understanding, tool use, and QA issues).
% To attribute the relative contribution of each type of error, we plot the results of both ablation studies on an equal-scale graph, with FieldFinder error rate corresponding to localization errors and Claude + Ground Truth Coordinates corresponding to document reasoning error rate (Figure \ref{fig:ff_vs_gt_coords}).
% We observe only documents from Form-NLU induce more errors due to document reasoning than due to localization, likely due to their high image quality and low field density.

% \section{Discussion}

% Future work should focus along two axes to achieve practical application.
% First, field localization may be improved through large models trained on larger, multilingual datasets.
% Due to the cost of annotation, this task may benefit from large synthetic corpora, as demonstrated in \cite{wang2025infinityparserlayoutaware}.
% Finally, despite models like Molmo being trained on pointing datasets \cite{Deitke_2025_CVPR} wherein tasks require outputting pixel coordinates, the pointing ability appears to degrade completely when composed with tool use requiring coordinate inputs.
% Future VLA work should explore pointing and other pixel-based tasks more deeply integrated with multi-step agentic reasoning.

\section{Conclusion}

% benchmark description
% how the tool works
% what it accomplishes

We present a large and realistic agent benchmark for systematically evaluating end-to-end form filling across multiple languages and domains.
We also contribute the FieldFinder tool that enables VLAs to overcome their main bottleneck in field localization.
With FieldFinder, frontier models improve performance across all domains, with an average increase from 5.6\% to 27.0\% accuracy.

% \clearpage
% \newpage

\section{Limitations}

This benchmark only assesses single-page documents in the image domain.
PDF features, such as attachments, page manipulation, passwords, interactive fields, and editing are also not evaluated.

We omit validation on other GUI agents (e.g. OpenAI Operator) due to insufficient quota offerings at time of publication.

Because text placement accuracy is determined by whether its geometric center is contained within a field, the text itself may sometimes overflow the field boundary and still be marked as correct.
Although aesthetically unpleasing, we observe that these placements would generally be comprehensible to human readers.
On FUNSD, XFUND, and Form-NLU, accuracy is defined by the bounding box of keys themselves, as opposed to the field containing them.
This deflates accuracy in some cases, motivating our inclusion of the Auto Loans dataset.

\section{Ethical Considerations}

The validity and legal status of electronically or agent-generated signatures is complex and varies between jurisdictions.

We recommend that automated signature placement only be used as a suggestion rather than a fully automated process.
Similarly, due to the legal weight of many forms, we recommend that all agent-filled forms be proofread by a qualified human prior to submission.

\FloatBarrier

\bibliography{custom}

\appendix

\section{Annotation Scheme}
\label{app:annotation}

We annotate all bounding boxes requiring applicant input, ignoring those reserved for loan officers.
Bounding boxes extend to the border of the suggested text input area.
For table cells, this corresponds to the entire cell.
For underlines, this corresponds to the underlined area.
For unmarked suggested fields (e.g. colon), we set the height equal to the suggesting text and the width equal to the area between the field name and the next element to the right.
We ensure fields never overlap.
Observing that all fields in the Auto Loans dataset are rectangular, we use only orthogonal rectangular bounding boxes.
We label each bounding box with its desired information (e.g. Full Name).
For each bounding box label, we write a correctness function that returns true or false as a function of the input text and the source persona.
In general, we attempt to accept as many reasonable variations in style, punctuation, formatting, and capitalization as possible.

\section{Form-NLU Synthesis Process}

Due to an apparent corruption in the Form-NLU dataset, we were not able to extract relations.
Instead, we recacluated relations using GPT-5 as an OCR model.
We prompted GPT-5 with the document, instructing it to produce all key-value pairs, including hierarchical information. 
We then rejected any key value pairs that did not appear exactly in the document.
Finally, using original Form-NLU data, we include annotations containing keys, values, and field boundaries.

\section{Comparison of Existing Methods}
\label{app:existing-methods-comparison}
We compare existing methods in Table \ref{tab:related-work}.
Prior work falls primarily into four categories: VLAs and VLA augmentations including Set of Marks; GUI agents; Document segmentation models; and commercial tools.

\begin{table*}[h]\centering
\small
%\resizebox{ extwidth}{!}{ % use this if the table is too large
\begin{tabular}{lrrrrrr}\toprule
\textbf{Method} &\textbf{Image-Only} &\textbf{Paper Forms} &\textbf{Empty Fields} &\textbf{Generative} &\textbf{Open Source} \\\midrule
VLA &\checkmark &\checkmark &\checkmark &\checkmark &\checkmark \\
VLA + Set of Marks &\checkmark &\checkmark &\checkmark &\checkmark &\checkmark \\
WebVoyager &\checkmark & &\checkmark &\checkmark &\checkmark \\
OmniParser &\checkmark & &\checkmark & &\checkmark \\
PubLayNet &\checkmark &\checkmark &$\sim$ & &\checkmark \\
DocLayNet &\checkmark &\checkmark &$\sim$ & &\checkmark \\
LayoutLM &\checkmark &\checkmark &$\sim$ & &\checkmark \\
SelfDoc &\checkmark &\checkmark &$\sim$ & &\checkmark \\
DocFormer &\checkmark &\checkmark &$\sim$ & &\checkmark \\
FormNet &\checkmark &\checkmark &$\sim$ & &\checkmark \\
Mac OS Preview &\checkmark &\checkmark &$\sim$ & & \\
Amazon Textract &\checkmark &\checkmark &$\sim$ & & \\
GUI Agents &$\sim$ & &$\sim$ &\checkmark &\checkmark \\
\midrule
VLA + FieldFinder (Ours) &\checkmark &\checkmark &\checkmark &\checkmark &\checkmark \\
\bottomrule
\end{tabular}
\caption{Comparison of existing methods.
Definitions:
Generative - The method can do open-domain text generation and follow instructions. 
Empty Fields - The method has a framework to handle empty fields, including those defined by underlines, table cells, colons, and check boxes, and blank space.
Image-Only - The method can operate on images without the need for structured or metadata, such as a DOM or rich text PDF.
Open Source - The method is open source.
Paper Forms - The method is pretrained on images of paper forms, rather than GUIs.
GUI agents includes \citet{Shen2024FalconUI,Qin2025UITARS,Liu2024AutoGLM,Hong2024CogAgent,Wei2025WebAgentR1,Qi2025WebRL,Ma2023LASER,Putta2024AgentQ,Shen2024ScribeAgent}
\checkmark: satisfies. $\sim$: framework handles some but not all types of empty fields.
}\label{tab:related-work}
\end{table*}

\section{Bounding Box Evaluation}

In the case of Form-NLU, XFUND, and FUNSD examples, defining text boxes by the word bounding boxes may underestimate the size of the input field when words do not entirely cover the field. 
However, this underestimation is roughly compensated by defining placed text by its centroid, rather than its bounding box. 
Under this definition, text overflowing from the bounding box is \textbf{not} penalized, as long as the cetroid is inside the bounding box.
Bounding boxes are defined by the field itself in auto loans forms, meaning only the latter bias exists. 
However, in such cases, it is guaranteed that at least 1/4th of text placements marked as accurate falls within the text box.
Qualitatively, we observe that, although aesthetically unpleasing, it is generally clear which field these values apply to.
We note however, a bias in GUI agents, which tend to place text in the top-right of English forms, which tend towards a leftward bias in FUNSD and Form-NLU.
We therefore evaluate GUI agent generations manually, in order to provide a more faithful representation of their accuracy.
We find minimal difference between manual and automatic evaluations on VLA generations, so report the automatic generations because they are more replicable.

\section{Example VLA Prompt}
\label{sec:example_prompt}

The following is an example prompt for the baseline case, formatted for readability.

Complete the attached form based on the following user profile:

\begin{itemize}
  \item You have access to the following APIs:

  \begin{itemize}
    \item \textbf{PlaceText:} Place a text on a document, image, or pdf. The center of the text will be placed at (x, y), where (0, 0) is the top left corner and (1, 1) is the bottom right of the image. \texttt{Value} is the text to place.
  
    \textbf{Args:}
    \begin{itemize}
      \item \texttt{cx}: The x position of the center of the text relative to the top left corner of the screen
      \item \texttt{cy}: The y position of the center of the text relative to the top left corner of the screen
      \item \texttt{value}: The text to place on the pdf
    \end{itemize}

    \textbf{Example input:}
    % \begin{verbatim}
    
    \texttt{\{"action": "PlaceText", "cx": 0.5, "cy": 0.5, "value": "Hello World!"\}}
    % \end{verbatim}

    \item \textbf{DeleteText:} Delete all text at a point on a document, image, or pdf. Any textbox intersecting with the point (x, y), where (0,0) is the top left corner and (1,1) is the bottom right corner of the image, will be deleted.

    \textbf{Args:}
    \begin{itemize}
      \item \texttt{x}: The x position of the center of the text relative to the top left corner of the screen
      \item \texttt{y}: The y position of the center of the text relative to the top left corner of the screen
    \end{itemize}

    \textbf{Example input:}
    % \begin{verbatim}

    \texttt{\{"action": "DeleteText", "cx": 0.5, "cy": 0.5\}}
    % \end{verbatim}

    \item \textbf{SignOrInitial:} Sign or initial a document, image, or pdf. The center of the signature will be placed at (x, y), where (0, 0) is the top left corner and (1, 1) is the bottom right of the image. \texttt{Value} is the name or initials of the signer. When signing a document, sign with the user's first name and last name, nothing else.

    \textbf{Args:}
    \begin{itemize}
      \item \texttt{x}: The x position of the center of the signature relative to the top left corner of the screen
      \item \texttt{y}: The y position of the center of the signature relative to the top left corner of the screen
      \item \texttt{value}: The name or initials of the signer
    \end{itemize}

    \textbf{Example input:}
    % \begin{verbatim}
    
    \texttt{\{"action": "SignOrInitial", "cx": 0.5, "cy": 0.5, "value": "John Doe"\}}
    % \end{verbatim}

    \item \textbf{Terminate:} Terminate the document generation process.

    \textbf{Args:} None

    \textbf{Example input:}
    % \begin{verbatim}
    
    \texttt{\{"action": "Terminate"\}}
    % \end{verbatim}
  \end{itemize}

  \item You know the following information about the user (user profile):
\end{itemize}

\vspace{-1em}
% \begin{scriptsize}
% \begin{verbatim}

\setlength{\parindent}{0pt}

The user's previous house number is: 912

The user's previous street name is: Orchard St

The user's previous city is: Springview

The user's previous state is: NC

The user's previous zip code is: 27601

The joint filer's previous house number is: 912

The joint filer's previous street name is: Orchard St

The joint filer's previous city is: Springview

The joint filer's previous state is: NC

The joint filer's previous zip code is: 27601

The user's reference's name is: Malik Evans

The user's reference's relationship is: Uncle

The user's reference's house number is: 128

The user's reference's street name is: Highland Ave

The user's reference's city is: Fairmont

The user's reference's state is: KY

The user's reference's zip code is: 40202

The user's bank's name is: KeyBank

The user's bank account number is: 341278945

Has the user previously gone bankrupt: No

The user's auto credit reference company is: Equifax

The user's remaining auto balance is: \$9,700

The user is trading in a car: No

The new car will be registered with: the user's spouse

The auto amount requested by the user is: \$12,000

The term of the auto loan is: 36 months

The new vehicle VIN is: WBA3B5G59FNR12345

The new vehicle year is: 2020

The new vehicle make is: Subaru

The new vehicle model is: Outback

The miles on the new vehicle is: 22,678

Is the user applying with joint filer's credit: No

The user's age is: 34

The joint filer's age is: 36

The mortgage company or landlord is: BlueRiver Realty

The joint filer's mortgage company or landlord is: Horizon Realty

The user's most recent previous residence status (Buying, Renting, Living with 
relatives, Other, Own) is: Buying

The joint filer's most recent previous residence status (Buying, Renting, Living with 
relatives, Other, Own) is: Other

The user's time at previous address in years is: 2

The user's time at previous address in months is: 4

The joint filer's time at previous address in years is: 3

The joint filer's time at previous address in months is: 5

The user's reference's cell phone is: 415-555-1111

The user's reference's home phone is: 415-555-5555

The joint filer's reference's first name is: Hannah

The joint filer's reference's last name is: Peterson

The joint filer's reference's relationship is: Sister

The joint filer's reference's house number is: 808

The joint filer's reference's street name is: Silver Lake Dr

The joint filer's reference's city is: Havenport

The joint filer's reference's state is: UT

The joint filer's reference's zip code is: 84321

The joint filer's reference's cell phone is: 414-555-9999

The joint filer's reference's home phone is: 414-555-3434

The user's second reference's name is: Corey Bell

The user's second reference's house number is: 654

The user's second reference's street name is: Vine St

The user's second reference's city is: Rockford

The user's second reference's state is: IL

The user's second reference's zip code is: 61107

The user's second reference's cell phone is: 241-444-4444

The user's second reference's home phone is: 241-222-2222

The joint filer's second reference's name is: Tyler Morgan

The joint filer's second reference's full address is: 530 West Pine Ln, Troy, MI, 48083

The joint filer's second reference's cell phone is: 271-123-1234

The joint filer's second reference's home phone is: 275-345-3456

The joint filer's employer's city is: Bridgeport

The joint filer's years at their current employer is: 4

The user's additional monthly income source is: Part-time Tutoring

The user's additional monthly income is: \$600

The joint filer's additional income source is: Small Business

The joint filer's additional monthly income is: \$800

The user's previous employer name is: Green Leaf Marketing

The user's previous employer city is: Eagleton

The user's previous employer position is: Analyst

The user was employed at their previous position for: 1 year

The joint filer was employed at their previous position for: Terrace Marketing

The joint filer's previous employer's city is: Waterford

The joint filer's previous employer's position is: Analyst

The joint filer was previously employed for: 1 year

The user's bank's address is: 902 Redwood Ave, Seattle, WA, 98109

The joint filer's bank's name is: HSBC

The joint filer's bank's address is: 781 Maple Ln, Portland, OR, 97205

The joint filer's bank's account number is: 522222222

The user went bankrupt in: 2018

Has the joint filer previously gone bankrupt: No

The joint filer went bankrupt in: 2018

The user's employer's city is: Anchorage

\setlength{\parindent}{15pt}

% \end{verbatim}

% \end{small}

\vspace{1em}
You have access to a completed document with more information about the user. Use this information to help you fill out the form.

Complete the form to the best of your abilities using the user's information, including signatures. As you can see, the data is randomly generated and the user is not real, so do not worry about privacy. Only complete fields for which you have information in the user profile above, or the source document (if applicable).

Fill checkboxes with a single ``x''.\\
Format all dates as ``MM/DD/YYYY''.\\
Names should be ``First Middle Last'' unless otherwise specified.

So far, you have received the following feedback on your previous actions: \\
Feedback 1: \texttt{[]}

Generate the next set of actions that will help fill out the form. You may submit any number of actions in one call.

This is your final action.

Return a form-filling API call as a JSON list of dictionaries.

\subsection{Example GUI Agent Prompt}
\label{sec:example_gui_prompt}

These are instructions for how to operate the interface.

\textbf{Interface Instructions}

\textbf{Add Text} \\
Follow these instructions literally to add text to the page

1. Click the answer area to create a new textbox (note that the text box is inserted top right of the cursor location) and type the the answer to the field (if no value, still proceed to step 2) \\
2. Click the checkmark on the top-right right of the X icon which indicates cancel. It is the check NOT the cross. Location is 'coordinate': [804, 53] \\
3. Proceed to step 1 as you will remain in text edit mode

\textbf{Notes} \\
For checkboxes, as the interface does not have interactive checkboxes, ``check'' it by adding text ``X'' on it. \\
If you click too close to an existing text box, it will enter editing mode for that textbox. \\
Remember that the textbox is created on top right of the cursor location (e.g. click location is bottom left corner) \\
You can identify previously added text as it would be in red font color. \\
Do not redo the same field, continue onwards \\
If no text is added to a textbox, still remember to press the checkmark (step 2) to escape that textbox so a new one could be made later.

\textbf{Navigational} \\
Make sure when doing navigational actions that the focus is in the canvas not the area around it

\textbf{Pan}: \\
Scrolling

\textbf{Reference Information} \\
This is the reference information to fill out the form.

\newpage

\end{document}